\documentclass[5p]{elsarticle2}

\usepackage{lineno,hyperref}

\usepackage[utf8]{inputenc} 
\usepackage[T1]{fontenc}    
\usepackage[tbtags,fixamsmath]{mathtools}

\usepackage{graphicx}
\usepackage{subcaption}
\usepackage{wrapfig}
\graphicspath{{resources/}}

\usepackage{array}
\usepackage{booktabs}       
\usepackage{multirow}
\usepackage{multicol}
\usepackage{adjustbox}
\usepackage{color}
\usepackage{soul}

\usepackage[inline]{enumitem}
\setlist[itemize]{leftmargin=*, noitemsep, topsep=0pt}
\setlist[enumerate]{leftmargin=*, noitemsep, topsep=0pt}

\usepackage{algorithm}
\usepackage{algpseudocode}
\include{macros}

\journal{Pattern Recognition}

\begin{document}

\begin{frontmatter}

\title{Key Protected Classification for Collaborative Learning}

\author[nle,bilkent]{Mert Bulent Sariyildiz~\fnref{fnbilkent}}
\ead{mertbulent.sariyildiz@naverlabs.com}
\fntext[fnbilkent]{Work done during the author was a master student at the Computer Engineering Department of Bilkent University.}

\author[metu]{Ramazan Gokberk Cinbis}
\ead{gcinbis@ceng.metu.edu.tr}

\author[bilkent,case-western]{Erman Ayday}
\ead{eayday@cs.bilkent.edu.tr}

\address[nle]{NAVER LABS Europe, France}
\address[bilkent]{Computer Engineering Department, Bilkent University, Turkey}
\address[metu]{Computer Engineering Department, Middle East Technical University, Turkey}
\address[case-western]{Department of Computer and Data Sciences, Case Western Reserve University, USA}

\begin{abstract}
Large-scale datasets play a fundamental role in training deep learning models.
However, dataset collection is difficult in domains that involve sensitive information.
Collaborative learning techniques provide a privacy-preserving solution, by enabling training over a number of private datasets that are not shared by their owners.
However, recently, it has been shown that the existing collaborative learning frameworks are vulnerable to an active adversary that runs a generative adversarial network (GAN) attack.
In this work, we propose a novel classification model that is resilient against such attacks by design.
More specifically, we introduce a key-based classification model and a principled training scheme that protects class scores by using class-specific private keys, which effectively hide the information necessary for a GAN attack.
We additionally show how to utilize high dimensional keys to improve the robustness against attacks without increasing the model complexity.
Our detailed experiments demonstrate the effectiveness of the proposed technique. 
Source code is available at \url{https://github.com/mbsariyildiz/key-protected-classification}.%
\fntext[copyright]{©2020. This manuscript version is made available under the CC-BY-NC-ND 4.0 license \url{http://creativecommons.org/licenses/by-nc-nd/4.0}.}

\end{abstract}

\begin{keyword}
    privacy-preserving machine learning, collaborative learning, classification, generative adversarial networks
\end{keyword}

\end{frontmatter}

\section{Introduction}

Deep neural networks have shown remarkable performance in numerous domains, including computer vision, speech recognition, language processing, and many more.
Most deep learning approaches rely on training over large-scale datasets and computational resources that makes the utilization of such datasets possible.
While large-scale public datasets, such as~ImageNet \cite{imagenet_cvpr09}, Celeb-1M \cite{guo2016msceleb}, and YouTube-8M \cite{abu2016youtube}, have a fundamental role in deep learning
research, it is typically difficult to collect a large-scale dataset for problems that involve processing of sensitive information. 
For instance, data privacy becomes a significant concern if one considers training models over personal messages, pictures or health records.

To enable training over large-scale datasets without compromising data privacy, decentralized training approaches, such as {\em collaborative learning framework} (CLF)~\cite{shokri2015privacy}, {\em federated learning}~\cite{mcmahan2017federated}, {\em personalized learning}~\cite{pmlr-v84-bellet18a,pmlr-v54-vanhaesebrouck17a} approaches have been proposed.
These training schemes enable multiple parties (private data holders) to train a single neural network model without sharing their sensitive, private data with each other.

In this paper, we consider improving privacy protection mechanism of CLFs. 
In a CLF, a target model is trained in a distributed way, where each {\em participant} contributes to training without sharing its (sensitive) data with other participants.
More specifically, each participant hosts only its own training examples, and a central server, called the {\em parameter server}, combines local model updates into a shared model.
Therefore, the training procedure effectively utilizes the data owned by all participants.
At the end, the final model parameters are shared with all participants.

However, there are cases where the original CLF approach fails to preserve data privacy due to the knowledge 
embedded in the final model parameters. In particular, \cite{fredrikson2015model} show that the parameters
of a neural network model trained on a dataset can be exploited to partially reconstruct the training 
examples in that dataset, which is called a {\em passive attack}. 
To mitigate this threat, one may consider partially corrupting the model 
parameters by adding noise into the final model \cite{chaudhuri2011differentially}.
The study by \cite{shokri2015privacy} also shows that differential 
privacy \cite{dwork2011differential} can be incorporated into CLF in a way that guarantees the indistinguishability 
of the participant data by perturbing parameter updates during training. 
However, such prevention mechanisms may introduce a difficult trade-off between classifier accuracy versus data
privacy level for training. 
Several other differential privacy based approaches, which introduce
noise injection methods \cite{phan2016differential,abadi2016dldp} or training frameworks 
\cite{papernot2016semi}, have recently been proposed.

It has been shown that collaborative learning approaches can also be vulnerable to {\em active attacks} 
\ie, training-time attacks~\cite{hitaj_17_ganattack}.
More specifically, a training participant can construct a generative adversarial
network (GAN) \cite{goodfellow2014generative} such that its GAN model learns to reconstruct training
examples of one of the other participants over the training iterations. For this purpose, the attacker 
defines a new class for the joint model, which acts as the GAN discriminator, and utilizes the samples 
generated by its GAN generator when locally updating the model. In this manner, the attacker effectively 
forces the victim to release more information about its samples, as the victim
tries to differentiate its own data from attacker class during its local model updates.
To the best of our knowledge no solution---other than introducing differential privacy to the CLF,
has previously been proposed against the GAN attack\footnote{
\cite{pmlr-v84-bellet18a} claims that their methodology can reduce the efficacy of the GAN attack, however leaves the analysis for a future work.}.

In this paper, we propose a novel classification model for collaborative learning that prevents GAN attacks by design. 
First, we observe that GAN attacks depend on the classification scores of the targeted classes. 
Based on this observation, we define a classification model where class scores are protected by class-specific keys, which we call {\em class keys}.
Our approach generates class keys independently within each training participant and keeps the keys private throughout the training procedure.
In this manner, we prevent the access of the adversary to the target classes, and therefore the GAN attack.
We also demonstrate that the dimensionality of the keys directly affect the security of the proposed model, much like the length of passwords.
We observe, however, that naively increasing the key dimensions can greatly increase the number of model parameters, and therefore, reduce the data efficiency and the classification accuracy.
We address this issue by introducing a {\em fixed neural network layer} that allows us to use much higher dimensional keys without increasing the model complexity. 
We experimentally validate that our approach prevents GAN attacks while providing effective collaborative learning on the MNIST \cite{lecun1998mnist} and Olivetti Faces \cite{samaria1994parameterisation} datasets, both of which are challenging datasets in the context of privacy attacks due to their relative simplicity, and therefore, the ease of reconstructing the samples in them.

To sum up, our contributions can be summarized as follows:
\begin{enumerate*}[label=(\Roman*)]
    \item We formalize a novel classification model for collaborative learning frameworks where we decouple the end-to-end classifier learning into 
          the shared representation learning and the private class prediction steps secured by class keys.
          Making the class predictions private within each participant enables us to prevent the GAN attacks,
          while learning a shared image embedding model which generalizes across the private data hosted by all participants.
    \item We derive a principled training formulation for collaboratively learning the proposed model when participants are allowed to access only 
          their own class keys.
    \item We show that high-dimensional keys can be used to improve the robustness against attacks and introduce the idea of using randomly generated fixed neural network layers to map image representations to higher-dimensional spaces without increasing the number of learnable parameters.
    \item We investigate the key-based classification setup that we propose for the purpose of supervised training where all the data is centralized.
          In this regard, we show that our regression-like loss formulation achieves comparable results with the discriminative cross-entropy loss on CIFAR-10/100 datasets~\cite{krizhevsky2009learning}.
\end{enumerate*}

The rest of the paper is organized as follows. 
In Section~\ref{sec:bg}, we present a detailed and technical summary of collaborative learning, GANs, and the GAN attack technique. 
In Section~\ref{sec:method}, we discuss the details of our approach. In Section~\ref{sec:experiments}, we provide an experimental validation of our model.
Finally, in Section~\ref{sec:conclusion}, we make concluding remarks.

\section{Related work}\label{sec:relwork}

Attacks against machine learning mechanisms and privacy preserving machine learning methods have become a popular research area over the recent years. 
In \cite{chakraborty2018adversarial}, authors discuss different types of adversarial attacks and countermeasures against them.
In this section, we describe the most relevant attacks and countermeasures to our work. 

We focus on reconstruction attacks, in which the goal of the adversary is to reconstruct the training samples of the other participants in a distributed learning setting. 
Fredrikson \etal show this threat via a passive attack, in which the adversary does not actively attack during the learning process, but it tries to reconstruct the training samples from the final model parameters~\cite{fredrikson2015model}. 
In \cite{melis19exploiting},  authors develop passive and active inference attacks to exploit the leakage of sensitive information during the learning process. 
In particular, they show the risk of membership inference and attribute inference about training data (\ie, infer properties that hold for a subset of the training data).
In a more recent work, Hitaj \etal~\cite{hitaj_17_ganattack} devise a powerful active attack against collaborative learning frameworks.
In this attack,  one of the participants in the collaborative learning framework is assumed to be an adversary. 
The adversary tries to exploit one of the classes belonging to other participants (victim) by using a generative adversarial network (GAN)~\cite{goodfellow2014generative}.
This attack is powerful in the sense that the adversary can actively influence the victim to release more details about its samples during the training process.
The GAN attack is aimed for global data distribution among all clients, therefore it is challenging to run this attack for specific clients.
Following this idea, Wang \etal~\cite{wang19beyong} propose another GAN based attack, in which there is a malicious server which leaks model updates of a particular participant (the victim), and a multi-task discriminator that leads GAN.
This way, the attacker may choose the victim intentionally.
By doing so, the authors show how to discriminate category, reality, and client identity of input samples.

Several countermeasures have been proposed to mitigate the attribute inference attacks for machine learning mechanisms. 
One line of research on this direction is based on the differential privacy concept. 
Differential privacy proposed by Dwork \etal~\cite{dwork2006differential} aims at providing sample indistinguishability with respect to the outputs of an algorithm (or a neural network model) when its input is slightly changed. 
This indistinguishability criterion is then used as a proxy measure to quantitatively evaluate how well the algorithm (or the model) can protect the privacy of a subject data.
This concept is formalized for empirical risk minimization in machine learning problems by Chaudhuri \etal~\cite{chaudhuri2009pplogreg,chaudhuri2011differentially}.
Song \etal and Rajkumar \etal apply differential privacy to stochastic gradient descent-based optimization problems~\cite{song2013sgd,song2015learning,rajkumar_12_multipartysgd}.
Following that, Shokri \etal, Abadi \etal, Phan \etal and Papernot \etal propose several ways of employing differential privacy for large-scale machine learning problems in the forms of (i) structured noise addition \cite{abadi2016dldp, shokri2015privacy, phan2016differential} and (ii) 2-step training methodology \cite{papernot2016semi}.
Using the differential privacy concept together with trusted hardware, in \cite{hynes2018efficient}, authors introduce a privacy-preserving deep learning framework called Myelin. 
Similar to differential privacy-based approaches, in \cite{zhang2018privacy}, to preserve the privacy of training samples, authors propose using an obfuscate function to add random noise (or new samples) to the training data before using it for model generation.
Although the use of differential privacy in machine learning problems has shown promising results, most of the differential privacy-based approaches offer a trade-off between the privacy and the utility of data, \eg increasing the protection reduces the utility of data, and hence the performance of the algorithm. 
Therefore, overall, privacy-preserving machine learning without recognition performance compromise remains as an unsolved problem.

Another line of research advocates the use of decentralized training schemes for privacy.
These strategies enable large-scale machine learning for scenarios, in which private datasets that contain sensitive information are hosted by multiple parties and therefore cannot be shared.
There are two main streams of approaches:
(i) multiple parties train a single model on the fly by means of contributing to the model updates individually using their private data~\cite{shokri2015privacy,mcmahan2017federated} and
(ii) multiple parties learn separate models over their private data, and then the final model is constructed by aggregating the information stored in the different models~\cite{hamm2016lpfmd,rajkumar_12_multipartysgd,pathak_multiparty_2010}.
Besides, any of the methods considered under this line can still adopt differential privacy or secure multi-party computation techniques \cite{mohassel17secureml,bonawitz17practical}, in which model updates can be encrypted before sharing them with others.

Other than these two major lines of research, the use of cryptographic tools has also been proposed for privacy-preserving machine learning~\cite{chakraborty2018adversarial,gonzalez2017training,gomezbarrero_2017_multibio,WANG20141321,ANEES2018289,CHEE2018273}. 
Several differential privacy-based solutions have been proposed to prevent the GAN attack, in which a noise signal is added to gradients during the learning phase in order to achieve differential privacy, and hence prevent the GAN attack \cite{xie2018differentially,xu2019_ganobfuscator}.
Different from these works, our proposed work provides a countermeasure against the GAN attack without requiring a trade-off for utility.

\begin{figure*}[t]
    \hspace{-0.5cm}
    \centering
    \begin{subfigure}{0.50\textwidth}
        \centering
        \includegraphics[height=6.4cm]{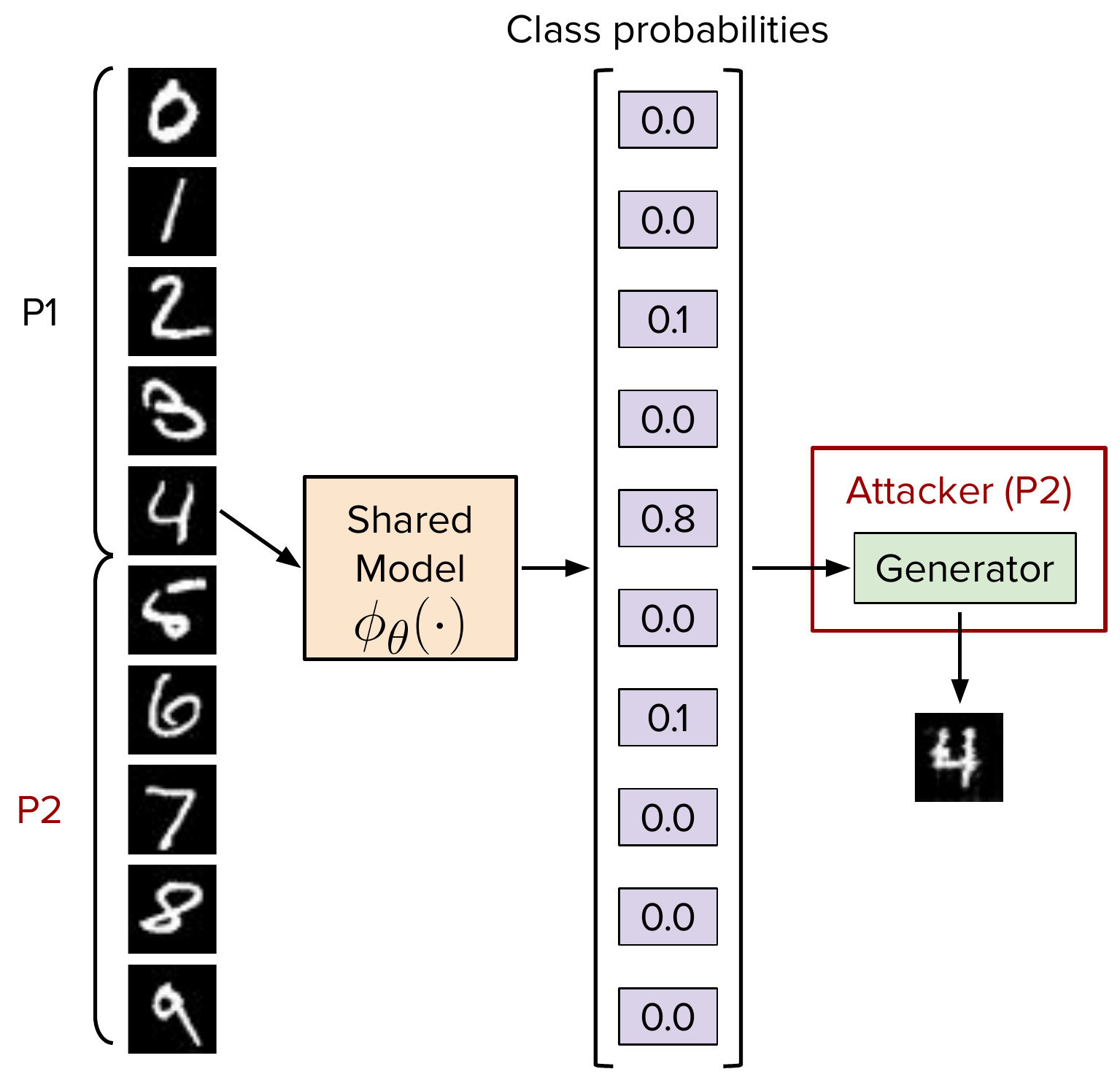}
        \caption{The compute chain in GAN attack.}
        \label{fig:overview-ganattack}
    \end{subfigure}%
    \begin{subfigure}{0.50\textwidth}
        \centering
        \includegraphics[height=6.4cm]{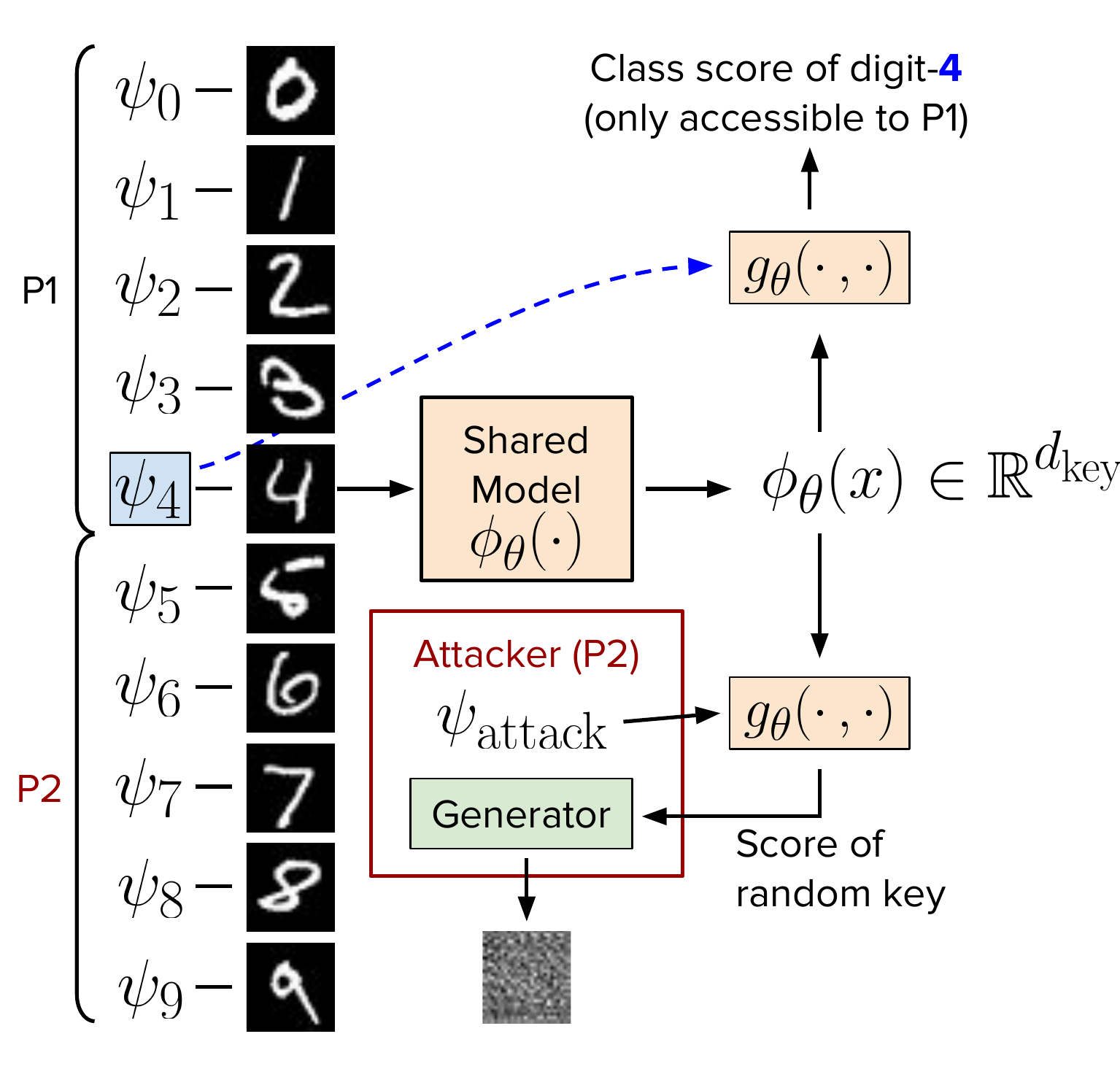}
        \caption{The compute chain in our proposed method.}
        \label{fig:overview-ours}
    \end{subfigure}
    \caption{The comparison of the compute chains constructed during
             the vanilla (left, a) and the modified key-protected (right, b) collaborative learning frameworks.
             In both scenarios, the dataset is split among two participants,
             P1 (an honest participant) and P2 (an adversary) and they train a shared neural network model
             by following the collaborative learning framework procedure defined in~\cite{shokri2015privacy}.
             In (a), the shared model directly outputs class prediction probabilities $\phi_\theta(x)$ for an image $x$ and this enables
             participants to train the model by optimizing a discriminative loss, \ie~cross entropy between the predicted class probabilities and 
             the ground truth labels of samples.
             However, outputting the class scores makes the shared model vulnerable to the powerful GAN attack~\cite{hitaj_17_ganattack}.
             In (b) the participants create a $d$-dimensional private class key $\psi_c$ for each of the classes for which they have training samples
             and the shared model outputs $d$-dimensional image embedding vector $\phi_\theta(x)$ for an image $x$.
             Moreover, instead of minimizing the cross entropy loss, they maximize the cosine similarity $f(\cdot \, , \cdot)$
             between the image embedding vector $\phi_\theta(x)$ and the private key $\psi_c$ of the correct class $c$.
             This way, the class probabilities are concealed from the attacker. Still, the adversary can attack by means of creating
             random class keys $\psi(\text{rand})$ (hoping that the random keys are close enough to any of the real ones created privately by the honest participant)
             yet it fails as it is unlikely that the random keys are close to the real ones.
             }
  \label{fig:overview}
\end{figure*}

\section{Background}\label{sec:bg}
Our work builds on the collaborative learning framework~\cite{shokri2015privacy} and 
tackles the generative adversarial network attack problem~\cite{hitaj_17_ganattack}.
Therefore, before providing the details of our approach, we first provide brief 
background on CLF and the GAN attack in the following.

\subsection{Collaborative Learning of Participants}\label{sec:bg-coll-learn}
The goal of CLF is to collaboratively train a shared model over private datasets of 
several participants such that the model generalizes across all the private datasets.  
For this purpose, CLF defines a protocol where each
participant shares with others information only about its learning progress, rather than the data directly, 
over the training iterations. Locally, participants train their model as usual 
using gradient based optimization, but share with others fractions of changes in model parameters, 
at predefined intervals. 
The framework is set up among participants based on the following components and the associated policies:
\begin{enumerate}[label=\Roman*.]
  \item A mechanism for participants to share parameter updates. This is typically realized by a trusted
        third-part parameter server (PS), using which the participants accumulate the model parameters
        by uploading and downloading fractions of their local parameter changes and central model parameters, respectively.
  \item A common objective and model architecture. All participants use the same model architecture and 
        training objective. Typically, participants declare class labels for which they have training data. 
  \item Meta-parameters. The hyper-parameters of the CLF setup, such as the parameter download fraction 
        ($\theta_d$) and the upload fraction ($\theta_u$), gradient clipping threshold ($\gamma$), 
        and the order of participants during training (e.g., round robin, random, 
                        or asynchronous) are typically predetermined~\cite{shokri2015privacy}.
\end{enumerate}
Once the framework is established, participants start training on their local datasets in a predetermined 
order. When a participant takes turn, it first downloads $\theta_d$ fraction of parameters from the 
parameter server and replaces them with its local parameters.
After performing several-steps training (for instance, one epoch of training)
on its local dataset, participant uploads $\theta_u$ fraction of resulting gradients to the parameter server.
It is also possible to incorporate differential privacy \ie, by injecting some form of noise to 
the uploaded gradients, to guarantee a certain level of sample indistinguishability for enhanced privacy protection.
But this procedure comes with a trade-off between the level of privacy and the efficiency (performance) of the accumulated
final model.
We encourage readers to refer to~\cite{shokri2015privacy} for the details of this learning protocol.

\subsection{Generative Adversarial Network}\label{sec:bg-gan}
Generative adversarial network~\cite{goodfellow2014generative} is an unsupervised learning process for 
learning a model of the underlying distribution of a sample set. 
A GAN model consists of two sub-models, called generator and discriminator. The generator corresponds to 
a function $\mathcal{G}(z;\theta_\mathcal{G})$ that aims to map each data point $z$ sampled from a prior distribution, 
\eg~uniform distribution $\mathcal{U}(-1,1)$, to a point $\hat{x}$ in the data space, 
where $\theta_\mathcal{G}$ represents
the generator model parameters. Similarly, the discriminator is a function $\mathcal{D}(x;\theta_{\mathcal{D}})$ that 
estimates the probability 
that a given $x$ is a real sample from the data distribution $p_{\mathcal{D}}$, 
where $\theta_\mathcal{D}$ represents the discriminator
model parameters. 

The generator and discriminator
models are trained in turns, by playing a two-player mini-max game. At each turn, the generator is 
updated towards generating samples that are indistinguishable from the real samples according to the current
discriminator's estimation:
\begin{equation}
    \min\limits_{\theta_\mathcal{G}} \,\, \mathbf{E}_{z \sim p(z)}\big[ \log(1 - %
        \mathcal{D}(\mathcal{G}(z;\theta_\mathcal{G}) ;\theta_\mathcal{D})) \big].
\label{eq:bg-gan-gen}
\end{equation}
The discriminator, on the other hand, is updated towards distinguishing the samples given by $\mathcal{G}$ from
the real ones:
\begin{equation}
\begin{split}
    \max\limits_{\theta_\mathcal{D}} \,\, & \mathbf{E}_{x \sim p(x)}\big[ \log(\mathcal{D}(x;\theta_\mathcal{D})) \big] + \\
                                          & \mathbf{E}_{z \sim p(z)}\big[ \log(1 - \mathcal{D}(\mathcal{G}(z;\theta_G));%
                                              \theta_\mathcal{D}) \big].
\label{eq:bg-gan-disc}
\end{split}
\end{equation}
GANs have successfully been utilized in numerous problems, \eg~see~\cite{Zhang_2017_ICCV, Yeh_2017_CVPR}.

\subsection{GAN Attack in Collaborative Learning}\label{sec:bg-attack}

~\cite{hitaj_17_ganattack} devise a powerful GAN-based active attack against
collaborative learning frameworks. In this scenario, an adversarial participant takes places during 
training in a collaborative learning framework setup in order to extract information about some class $c_\text{attack}$
for which any of the other honest participants has samples\footnote{The adversary may additionally 
have its own real classes and a real dataset, but for the sake of simplicity, we assume that the 
adversary works only on its privacy attack.}.
To do that the attacker utilizes the shared model as a discriminator network
and trains a generator network to capture the data manifold of the class $c_\text{attack}$.
Besides, the attacker announces an incorrect, unique class $c_\text{fake}$ to label 
the examples sampled from the generator network.
An overview of this attack is depicted in Figure~\ref{fig:overview-ganattack}.

In~\cite{hitaj_17_ganattack} it is shown that such an approach effectively turns collaborative learning framework into a GAN training setup
where an adversary takes the following steps:
\begin{enumerate}[label=\Roman*.]
  \item The adversary updates its generator network towards producing samples that are classified as 
        class $c_\text{attack}$, by the shared classification model. 
  \item The adversary takes samples from its generator, labels them as $c_\text{fake}$ and updates the 
        shared classification model towards
        classifying the synthetic samples as class $c_\text{fake}$. 
\end{enumerate}
There are two key aspects of the GAN attack. First, throughout the training iterations, the adversary continuously
updates its generator, therefore, it can progressively improve its generative model and the
reconstructions that it provides. Second, since the adversary defines the class $c_\text{fake}$ as part of 
the shared model, the participant that hosts $c_\text{attack}$ updates the shared model towards minimizing
the misclassification of its training examples into class $c_\text{fake}$. Over the iterations, 
this practically forces the victim into releasing more detailed information
about the class $c_\text{attack}$ while updating the shared model~\cite{hitaj_17_ganattack}.
This latter step makes the GAN attack particularly powerful as it influences the training of all 
participants, and, it is also the main reason why the technique is considered as an active attack.

It is also worth mentioning the similarities between the training objectives of the generator networks in the GAN-based attack and in~\cite{NIPS2016_6125}.
In the GAN-based attack scenario, the adversary trains from scratch a local generator while all participants (including the attacker) train from scratch a shared classifier.
As the class logits produced by the classifier is accessible to all participants, during training, the attacker utilizes this classifier as a discriminator network much like Salimans \etal~\cite{NIPS2016_6125} train generator networks.
However, different from the case in~\cite{NIPS2016_6125}, in GAN attacks there is no separate class in the shared classifier to denote if a sample is ``real'' or ``fake''.
Instead, first the attacker decides on a class to attack ($c_\text{attack}$), then it uses the shared classifier as a reference to train its local generator in a way that the generated samples (fake samples) maximize the probability of class $c_\text{attack}$ according to the shared model.
In this sense, the shared classifier acts as a discriminator. 
Besides, in their experimental setups, Hitaj \etal~\cite{hitaj_17_ganattack} let the attacker to train its generator while the participants train the shared classifier.
Therefore, the optimization dynamics are also similar to that of GANs, \eg, early in training, the gradients back propagating from the classifier are in good shape.%

In this paper, privacy model and proposed protection mechanism completely follow 
the threat model defined in~\cite{hitaj_17_ganattack}. In our experiments, we simulate the 
experimental setups of~\cite{hitaj_17_ganattack,shokri2015privacy} without introducing any extra assumptions.

\section{Proposed Method}\label{sec:method}

In this section, we describe the details of our proposed approach. 
In Section~\ref{sec:method:keyprotected},
we present our class key-protected classification model, as a prevention mechanism against the GAN attack~\cite{hitaj_17_ganattack}.
In Section~\ref{sec:method:training}, we show
how such a model can be trained in a distributed manner when each participant has access only to the keys of its own classes,
while leveraging the fundamental tools given the distributed learning framework of Shokri \etal~\cite{shokri2015privacy}.
In Section~\ref{sec:method:keygeneration}, we discuss practical considerations in generating
random class keys.
In Section~\ref{sec:method:highdimensions}, 
we propose an extension of our approach that
enables efficient incorporation of high dimensional keys towards minimizing the risk
of a successful GAN attack within our key-protected classification framework.
Finally, in Section~\ref{sec:method:summary}, we present a summary of the proposed approach.

\subsection{Key Protected Classification Model}\label{sec:method:keyprotected}

Our goal is to train a deep (convolutional) neural network based multi-class classifier in a collaborative manner.
Our starting point is the observation that the GAN attack relies on the knowledge of classification
scores of samples belonging to the target class $c_\text{attack}$ throughout the training iterations, 
as discussed in Section~\ref{sec:bg-attack} and in Figure~\ref{fig:overview-ganattack}.
To prevent the attack in a collaborative learning setup, 
we aim to mathematically prevent each participant from estimating 
the classification scores for the classes hosted by the other participants.  
For this purpose, we introduce 
class-specific keys for all classes and parameterize the classification function in terms of 
these keys in a way that makes classification score estimation without keys
practically improbable. 

In our approach, we require each participant to generate random class keys for its classes 
during initialization, and keep it private until the end of the training process. 
We denote the key for class $c$ by $\psi_c \in \mathbb{R}^{d_\text{key}}$, where $d_\text{key}$ is the 
predetermined dimensionality of each key. In order to protect the model using class keys,
we first define 
the network with model parameters $\theta$ as an embedding function 
$\phi_{\theta}(x)$ that maps each given input $x$, \eg~an image, from the source domain to a 
$d_\text{key}$-dimensional $\ell_2$-normalized vector. Then,  we define the classification score for class $c$ by a simple
dot product between the embedding output and the class key: 
\[g_\theta(x,\psi_c) = \langle \phi_{\theta}(x), \psi_c \rangle,\]\label{eq:scoringfunc}
where $\psi_c, \phi_{\theta}(x) \in \mathbb{R}^{d_\text{key}}$. 
We note that the class keys are analogous to classification weight vectors, or equivalently, the components of the
last fully-connected layer of a standard (mainstream) feed-forward classification neural network.
However, unlike the case in a standard neural network model where the classification weights vectors are
learned during training, here, these vectors are pre-generated and kept fixed throughout the training process. 
The training, therefore, takes the form of learning the embedding model $\phi_{\theta}(x)$. 

Therefore, in contrast to a standard classification model where the model produces all class scores, in the proposed
formulation, a participant can compute class scores only for the classes for which it has class keys. As a result, a
participant does not have access to the class scores that are necessary by definition for the GAN attack.
The overview of our proposed approach is given in Figure~\ref{fig:overview-ours}.

Once the embedding model is learned and training is completed, 
all class
keys are made public so that the resulting model can be used to make predictions for all classes.
The final classification function takes the form of choosing the class whose key leads to the highest classification score:
\begin{equation}
\arg\max\limits_{c\in C^\text{all}} \,\, \langle \psi_c, \phi_{\theta}(x) \rangle, 
\label{eq:prediction}
\end{equation}
where $C^\text{all}$ is the set of all classes over all participants.

To ensure that the embedding function $\phi_{\theta}(x)$ provides 
normalized embedding vectors as required by the definition, an $\ell_2$-normalization layer is appended
to the corresponding unnormalized embedding network $\phi_{\theta}^{u}(x)$, so that:
\begin{equation}
    \phi_{\theta}(x) = \frac{\phi_{\theta}^{u}(x)}{\| \phi_{\theta}^{u}(x) \|}. \label{eq:phidef}
\end{equation}
A clear reason for incorporating $\ell_2$-normalization surfaces naturally in the derivation of our training
formulation, which we explain in the next subsection.

Finally, we note that the method is presented with the assumption that only a single adversary exists, for the sake of brevity.
Our framework, however, naturally handles multiple attackers without requiring any modifications, and, in fact, in Section~\ref{sec:experiments},
we do present experimental results for multiple attackers.

\subsection{Learning with Restricted Class Key Access}\label{sec:method:training}

In this section, we describe our approach for training the classification model, by making updates 
locally at each participant, using the restricted training set 
owned by the participants. Suppose that a participant owns $n$ training examples, 
represented by a set of tuples ${(x_i,c_i)}_{i=1}^n$. 
The goal of the local model update is to update the embedding model $\phi$ such that it 
minimizes the negative label likelihood of the training examples by regularized risk minimization:
\begin{equation}
    \min_\theta -\sum_{i=1}^n \log p_\theta(c_i|x_i) + R(\theta) \label{eq:trainobj1},
\end{equation}
where $R(\theta)$ is the regularization function. 

We aim to obtain label likelihoods $p_\theta(c|x)$ based on the scoring function $g_\theta(x,\psi)$. However, by design,
a participant is not allowed to access keys $\psi_c$ of other classes, therefore, cannot estimate the class
probability distribution via a conventional softmax operator over the set of unnormalized class scores.
Therefore, in order to define label likelihoods, we generalize the softmax operator by re-defining it as the ratio of exponentiated target
class score and the expectation of exponentiated class scores over all possible class keys:
\begin{equation}
    p(c|x) = \frac{\exp g_\theta(x,\psi_c)}{\mathbb{E}_{\psi}[\exp g_\theta(x,\psi) ]}.
\end{equation}
One way to interpret this definition is applying softmax over infinitely many classes, where class keys follow some
probability distribution. Assuming that class keys are obtained by sampling from the $d_\text{key}$-dimensional standard normal
distribution, the expectation in the denominator can be re-written as:
\begin{equation}
    \mathbb{E}_{\psi}[ \exp( g_\theta(x,\psi) ) ] = \int f(\psi) \exp g_\theta(x,\psi) d\psi, \label{eq:expsoftmax}
\end{equation}
where $f(\psi)$ is the probability density function for standard multivariate normal distribution. It can be shown that
the expectation yields the value $\exp(0.5 \| \phi(x) \|^2)$ (See~\ref{appx:exppsi}). By plugging this result
into Eq.~\ref{eq:trainobj1} and re-arranging the terms, we obtain:
\begin{equation}
    \min_\theta -\sum_{i=1}^n g_\theta(x_i,\psi_{c_i}) + 0.5 \sum_{i=1}^n \| \phi_\theta(x_i) \|^2  + R(\theta).
    \label{eq:trainobj2}
\end{equation}
Here, the first term corresponds to maximizing the classification scores, \ie~the sum of inner product values between each pair of sample embedding $\phi(x)$ 
and the corresponding class key $\psi_{c}$. The second term applies $\ell_2$ regularization over the $\phi(x)$
embeddings, which limits the scale of the vectors
produced by the embedding network $\phi(x)$. Finally, the third term is simple regularization over the
model parameters, for which use $\ell_2$ regularization.

The second term in Eq.~\ref{eq:trainobj2} is worth discussing in a bit more detail: unlike conventional $\ell_2$
regularization over the network parameters (\ie~$R(\theta)$), it applies $\ell_2$ regularization to the outputs of the
embedding network. This term, therefore, can be interpreted as a way to prevent learning a degenerate embedding
model that reduces the objective function by making the embedding vectors, and therefore the classification scores,
excessively large. It is also an interesting result in the sense that it appears naturally in our derivation. We note
that its weight ($0.5$) could be altered by using a different temperature constant in Eq.~\ref{eq:expsoftmax}, which
could be another hyper-parameter. However, as denoted before, we opt to fix the scale of the embedding vectors by incorporating
a final $\ell_2$ normalization layer into the network $\phi(x)$, which forces the network to focus on the directions,
and not the scales, of the embedding vectors. This is our primary motivation in incorporating the $\ell_2$-normalization
layer, and, it converts the second term into a constant value of $0.5 n$, which we drop from the objective function.

By plugging in the definitions of all terms into Eq.~\ref{eq:trainobj2} and dropping the second term from Eq.~\ref{eq:trainobj2}, we obtain the final form of the objective function:
\begin{equation}
    \min_\theta -\sum_{i=1}^n \langle \phi_{\theta}(x_i), \psi_{c_i} \rangle + \lambda \| \theta \|^2,
    \label{eq:trainobj3}
\end{equation}
where $\lambda$ is the regularization weight.
The final formulation can be interpreted as a regression-like training approach that aims to maximize the expected correlation
across the sample embeddings and the corresponding class keys.

\subsection{Class Key Generation}\label{sec:method:keygeneration}

In our approach, even if an adversary gets involved during training, it cannot target a 
particular class without knowing its class key. However, there is still a chance that 
the adversary may target an arbitrary class by using a randomly generated key $\psi_\text{attack}$ as the 
target key, and, aim to reconstruct the samples belonging to one of the classes without 
necessarily knowing the identity of the targeted class. In principle, such an attack can be 
successful if the randomly generated key is sufficiently similar to one of the actual 
class keys, if \ie~$ \| \psi_c - \psi_\text{attack}\|_2 \le \delta$ for any $c$.
However, we emphasize that there is no supervisory signal that the attacker can 
utilize to guess private class keys better than random. It is also not possible to run a 
GAN attack to reconstruct class keys, due to the lack of any reference score that can be
leveraged for this purpose.

To protect the system against such random key-based brute-force attack, it is essential to reduce the probability of approximately 
replicating class keys by randomly generating keys. 
Therefore, we need to minimize the probability of generating keys that will lead to scores 
highly correlated with one of the class scores, without relying on the restrictions on the 
keys used by the participants, such that an adversary will (most likely) not be successful 
in training a generative model through a GAN attack.

\begin{figure}
  \centering
  \includegraphics[width=4.7cm]{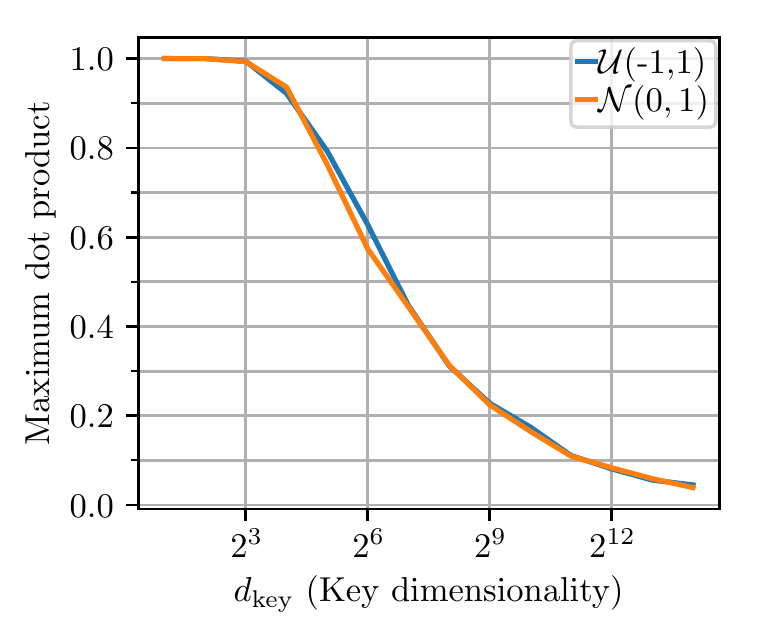}
  \caption{The correlation between two random vectors with respect to their dimension $d$.
           For $d_\text{key} = 2^1,\cdots,2^{14} $ we randomly generate 100 $\ell_2$-normalized vectors 
           by sampling either from $\mathcal{N}(0,1)$ or $\,\mathcal{U}(-1,1)$. 
           Then, we find maximum of pair-wise dot products of these vectors. 
           We repeat this process 1000 times for each $d_\text{key}$ and plot the maximum of 
           maximums of the pair-wise dot products.}
  \label{fig:exp-key-selection}
\end{figure}

One may consider addressing this problem by determining all class keys in a centralized manner. However, such a
prevention technique is not reliable as it is typically not possible to enforce participants to use the assigned keys,
\ie~the adversary may still attempt to attack with a random key. Also it is possible that the central server might be compromised.
For these reasons, we let each participant to generate its class keys independently.
To further reduce the probability of key ``conflicts'', \ie~having highly-correlated class key pairs,
we opt to using high-dimensional normally-distributed class keys and apply $\ell_2$-normalization to them in practice.
Using $\ell_2$-normalized keys avoids potential biases across classes that can be caused by 
differences in class key scales.
We also observe that as the key dimensionality increases, $\ell_2$ normalized keys tend to be
progressively less correlated. 
This observation can easily be verified in an empirical manner:
in Figure~\ref{fig:exp-key-selection}, we randomly generate 100 $\ell_2$-normalized vectors,
find the maximum of pair-wise dot products of these vectors and plot the maximum of 
maximums of the pair-wise dot products by repeating this process 1000 times for a variety of $d_\text{key}$ values.
From the figure, we can see that as the size of keys increase, the maximum overlap across sampled key pairs sharply diminishes.
From this empirical finding, we can claim that it is more robust to generate 
high dimensional vectors to prevent GAN attacks. 

We note that while our decentralized scheme addresses the key generation problem to a large-extend,
it may possibly lead to complications in two ways. First, if the participants have overlapping classes,
they will almost certainly generate and use completely different keys for their shared classes.
Fortunately, it turns out that this does not necessarily lead to the failure of the framework
and the approach behaves well to a large extend, which we experimentally verify in Section~\ref{sec:experiments}.

Second, the use of very high-dimensional keys may lead to training difficulties due to a drastic increase in
the model complexity. We show how to address this problem in a technical and principled way in the
following subsection.

\subsection{Learning with High Dimensional Keys}\label{sec:method:highdimensions}

In the proposed framework, as also previously discussed, it is important to have distinctive class keys,
with minimal cross-class key correlations for both the quality of the resulting classifier
and the success of the GAN-attack prevention mechanism (against the random-key attacks).
In order to reduce the probability of having highly-correlated pairs of 
randomly and independently generated class keys, we aim to use very high dimensional
vectors. However, naively increasing the class key dimensionality 
undesirably increases the number of trainable parameters in the last layer of the $\phi_\theta(x)$ network.
Therefore, this increase in the dimensionality of key vectors also increases the 
complexity of overall model architecture which may 
\begin{enumerate*}[label= (\Roman*)]
  \item slow down training significantly,
  \item cause over-fitting to training samples,
  \item and therefore, lead to a poor test performance.
\end{enumerate*}

To overcome these problems, we propose to add a \textit{fixed} dense layer with an activation function
at the last layer of the architecture. By doing so, parameters of this 
dense layer are stochastically predetermined and kept unchanged throughout training. Thus, the layer does not 
impose any extra trainable parameters to be learned. This layer, therefore, effectively maps the preceding low-dimensional embedding vectors
to a much higher dimension space.
We pre-define the parameters of this layer and keep it shared among all participants.
We randomly initialize the parameters of this layer and share the layer among all participants.
During the collaborative training phase, the layer is kept frozen.

\begin{algorithm}[t]
  \caption{Key Protected Classification for Collaborative Learning}
  \label{alg:pseudo}
  \begin{algorithmic}[1]
    \renewcommand{\algorithmicrequire}{\textbf{Pre-Training Phase:}}
    \renewcommand{\algorithmicensure}{\textbf{Training Phase}}
    \Require
    \State Participants agree on the collaborative learning framework parameters (see Section~\ref{sec:bg-coll-learn}) and also the dimensionality of private class keys $d_\text{key}$.
    \State Each participant $p$ generates a random class key $\psi_c^p$ for each class $c$ of its classes.
    \State Each attacker participant $p$ generates (at least) one extra placeholder $c_\text{fake}^\text{p}$ and the corresponding class key $\psi_\text{attack}^\text{p} $ for running a GAN attack towards some random class key through adversarial training.
    \Ensure
    \For{epoch = $1$ to $n_{epochs}$}
        \For{$p$ in Participants}
            \State download $\theta_d$ fraction of parameters from PS
            \State replace downloaded parameters with local ones
                \If {$p$ is an attacker}
            \State train the local generator by using $\psi_\text{attack}^\text{p}$
            \State generate $M$ samples from generator
            \State label the generated samples as $c_\text{fake}^\text{p}$
            \State merge generated samples with local dataset
            \EndIf
            \State train local model with local dataset
            \State upload $\theta_u$ fraction of differences in parameters to PS
        \EndFor
    \EndFor
      \State Participants optionally publish the class keys and labels they host.
  \end{algorithmic}
\end{algorithm}

We experimentally verify the effectiveness of this approach:
in Section~\ref{sec:exp-coll-learn-eval} and Section~\ref{sec:exp-preventing-gan-attack}
we show that key dimensionality can be increased without deteriorating the participant training convergence using the proposed fixed-mapping layer and 
as a result, the random-key GAN attacks can be prevented.

\subsection{Summary}\label{sec:method:summary}

To summarize the overall framework, we present the list of main steps in Algorithm~\ref{alg:pseudo}.
We note that the underlying collaborative training scheme is the same as the vanilla collaborative
learning algorithm of Shokri \etal~\cite{shokri2015privacy}. However, unlike~\cite{shokri2015privacy}, here the participants 
and the adversary additionally create random private class keys, and, train the proposed key-protected classification scheme
instead of a traditional classifier. 

Finally, we note that publishing class keys and labels is listed as an optional step: if the goal is to collaboratively
train a deep network only for representation learning, the class keys and labels may be left private even after training. In this case,
each participant may use the learned representation as needed. However, if the goal is to collaboratively
train a full classification model, then clearly both the keys and the class labels need to be published by the
participants.

\section{Experiments}\label{sec:experiments}

In this section we demonstrate that our proposed solution prevents GAN attacks while 
enabling effective collaborative classifier learning over the participants. We first explain the experimental
setup in Section~\ref{sec:expsetup}. We verify that our key-based learning formulation 
performs well in absence of an adversary in Section~\ref{sec:exp-coll-learn-eval}. 
We empirically demonstrate that our proposed approach prevents GAN attacks
in Section~\ref{sec:exp-preventing-gan-attack}. We evaluate the performance of the collaborative learning approach when
there are overlapping classes across the participants in Section~\ref{sec:sharedexp}. And finally, we demonstrate that 
key-based regression yields comparable results with cross entropy in Section~\ref{sec:key-versus-xent}.

\subsection{Experimental setup}\label{sec:expsetup}

We perform experiments on the well-known MNIST handwritten digits~\cite{lecun1998mnist} and AT\&T Olivetti Faces~\cite{samaria1994parameterisation}
datasets. We choose these datasets as they provide a particularly challenging and suitable experimental setup for our purposes:
\begin{enumerate*}[label= (\Roman*)]
  \item Both of these 
  datasets contain samples that are relatively easy to generate, therefore,
  they are particularly challenging to protect against a GAN attack. 
  While several improvements have been proposed for improving GANs, \eg~\cite{NIPS2016_6125, mirza2014conditional,
  NIPS2017_7126, che2016mode}, it is well-known that GANs can be difficult to train~\cite{goodfellow2016nips}. 
  Therefore, we want use datasets where the GAN-attack generator can easily capture relevant data statistics and perform a successful GAN attack.
  This situation allows us to evaluate the proposed CLF scheme in a challenging scenario.
  \item Qualifying the reconstructions obtained by the adversary is relatively 
  easy on these datasets, which allows us to more confidently argue about the success of our formulation.
\end{enumerate*}

\begin{figure*}[t]
    \centering
    \begin{subfigure}{0.40\textwidth}
        \centering
        \includegraphics[height=4cm,trim={0 0.25cm 0 0},clip]{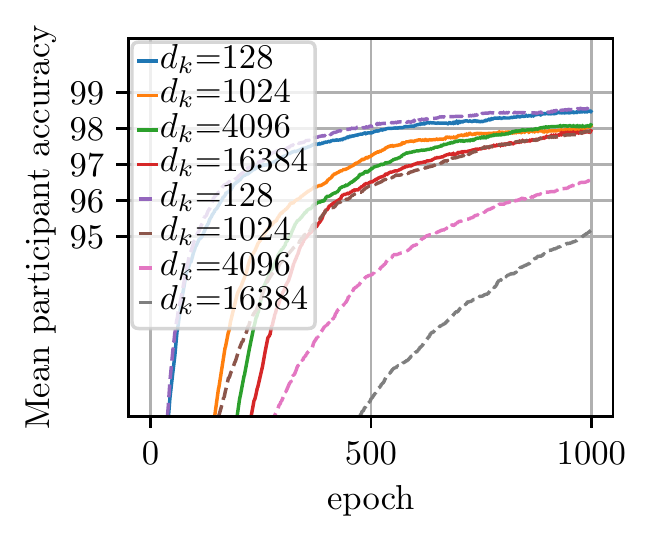}
        \caption{MNIST with 2 participants.}
    \end{subfigure}~
    \begin{subfigure}{0.60\textwidth}
        \centering
        \includegraphics[height=4cm,trim={0 0.25cm 0 0},clip]{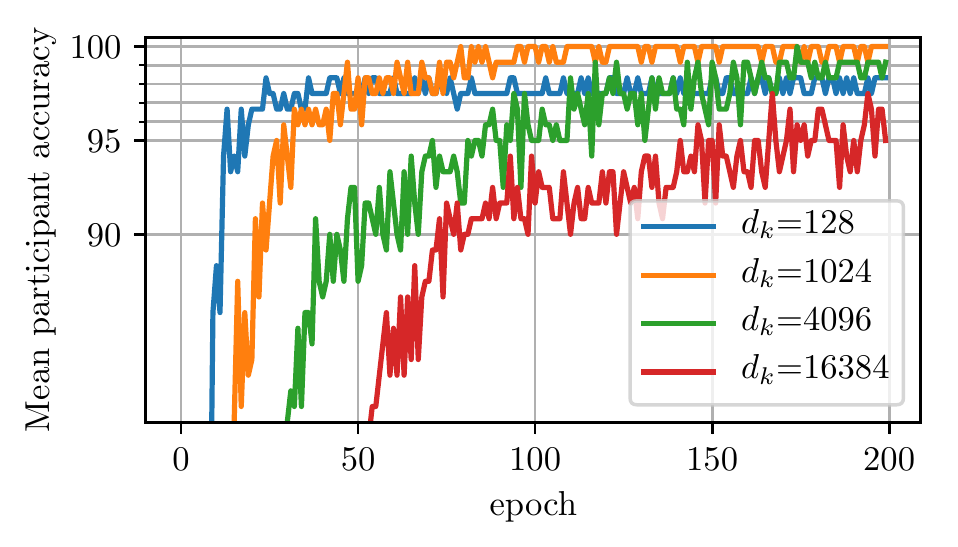}
        \caption{Olivetti Faces with 2 participants.}
    \end{subfigure}\\
    \begin{subfigure}{0.40\textwidth}
        \centering
        \includegraphics[height=4cm,trim={0 0.25cm 0 0},clip]{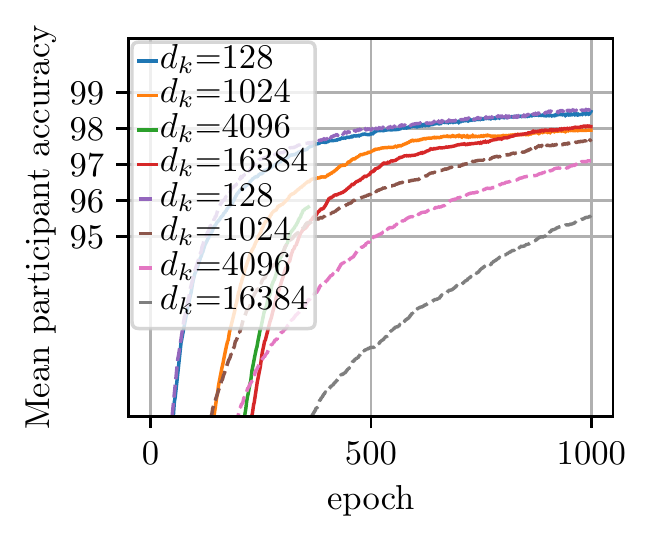}
        \caption{MNIST with 3 participants.}
    \end{subfigure}~
    \begin{subfigure}{0.60\textwidth}
        \centering
        \includegraphics[height=4cm,trim={0 0.25cm 0 0},clip]{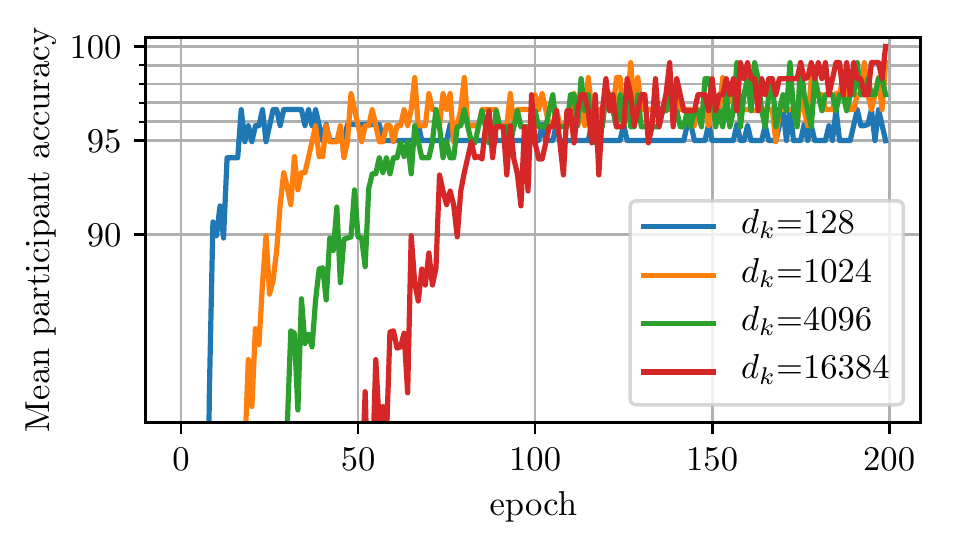}
        \caption{Olivetti Faces with 3 participants.}
    \end{subfigure}\\
    \begin{subfigure}{0.40\textwidth}
        \centering
        \includegraphics[height=4cm,trim={0 0.25cm 0 0},clip]{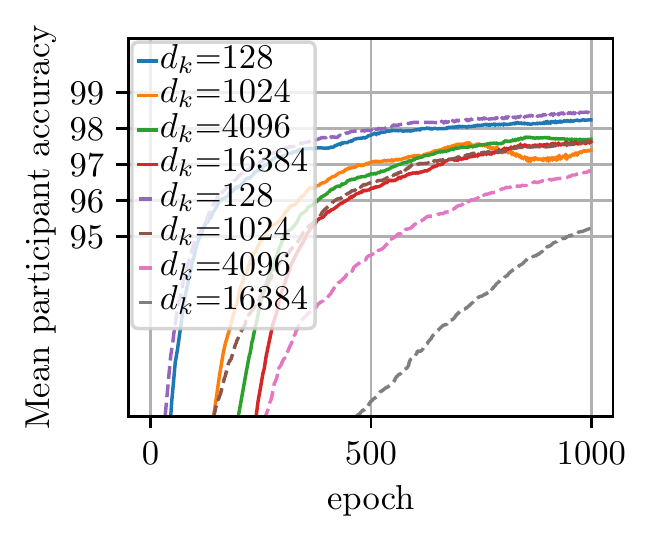}
        \caption{MNIST with 5 participants.}
    \end{subfigure}~
    \begin{subfigure}{0.60\textwidth}
        \centering
        \includegraphics[height=4cm,trim={0 0.25cm 0 0},clip]{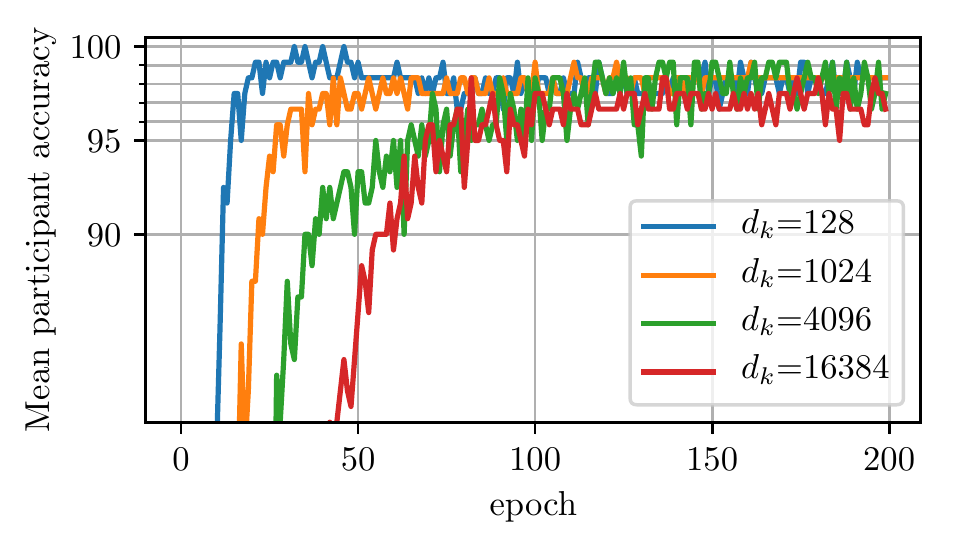}
        \caption{Olivetti Faces with 5 participants.}
    \end{subfigure}
    \caption{Mean participant accuracies obtained in collaborative 
            learning with 2, 3, and 5 participants over the MNIST and the Olivetti Faces
            datasets. The dashed lines in the MNIST figures indicate that there are fixed layers in the
            local models of participants
            (see Section~\ref{sec:method:highdimensions} for details).
            We see that using a fixed layer 
            tends to delay the training convergence 
            with similar final accuracy.
            In the Olivetti Faces plots, fixed layer based experiments are excluded for brevity.}
  \label{fig:exp-coll-learn}
\end{figure*}

We observe that collaborative learning with 5 or more participants is challenging as the parameters in the PS tend to overfit quickly to the local datasets of participants.
To overcome this, we use stochastic gradient descent (SGD) to optimize classifier and generator networks.
Classifiers are composed of convolutional and fully-connected layers with LeakyReLU~\cite{maas2013rectifier} non-linearity.
The generator architecture consists of convolutional layers with batch normalization~\cite{ioffe2015batch} and ReLU non-linearity.
We find that learning with fixed layer is compelling especially when the fixed layer transforms embeddings to a very high dimensional space (\eg, to $\mathbb{R}^{16384}$). 
We address this problem by using layer normalization~\cite{ba2016layer} after the fixed layer. 
This additional normalization introduces only $2 \times d_{\text{key}}$ parameters to train.
We carefully tune learning rate and $\lambda$ on validation sets.
For meta-parameters of collaborative learning framework, we set $\theta_d$ and $\theta_u$ to $1.0$ since~\cite{hitaj_17_ganattack} shows that the GAN attack also works for smaller values of $\theta_d$ and $\theta_u$.
We also exclude gradient selection mechanism, $\gamma$ and $\tau$ from the experiments. 
Finally, we note that increasing the key dimensionality can increase the robustness of the classification model against the GAN attack.
However, setting it to a too large value may lead to numerical instabilities and/or delay the convergence.
Therefore, $d_{\text{key}}$ should be treated as an hyper-parameter and tuned specifically for each {\em model architecture} on a separate held out public dataset with relevant content.
The experiments are implemented in TensorFlow~\cite{tensorflow2015-whitepaper} and PyTorch~\cite{NEURIPS2019_9015}.

\subsection{Collaborative Learning Evaluation}\label{sec:exp-coll-learn-eval}

In this section, we evaluate our collaborative learning framework formulation with private class keys, in the absence of an adversary,
in order to show that our formulation enables effectively learning a shared model. 
For this purpose, we examine how the key size $d_{\text{key}}$ and a fixed layer affect the 
test set accuracy of the local models, over the training iterations.
To do that, we split MNIST among $2$, $3$ and $5$ participants, run two different collaborative learning frameworks
(one with fixed layers and one without fixed layers) for $d_{\text{key}} \in \{128, 1024, 4096, 16384\}$
and report mean participant accuracies (MPA) during the training.
We also examine mean participant accuracies with respect to $d_{\text{key}} \in \{128, 1024, 4096, 16384\}$ and 
number of participants in a collaborative learning framework when there is no fixed layer in local models of participants
for the Olivetti Faces dataset. 
The results of these experiments are given in Figure~\ref{fig:exp-coll-learn}.  

{\bf Observations.}
We find that even on MNIST it takes considerably longer time to achieve a high classification accuracy
compared to the centralized case when applying collaborative learning framework in its vanilla form,
\eg~without any ``hacks'' to stabilize and speed up learning. We believe that this is due to the noise introduced
by participants when they upload parameters to the server from their local models after 
performing one epoch of local training. 
We note that, however, in this work we focus not on
improving the training performance of collaborative learning framework, 
but instead on evaluating the proposed approach within the existing CLF protocol.

MNIST experiments show that fixed layers yield better scores only when $d_{\text{key}} = 128$.
For other key dimensionalities and when same hyper-parameters used,
we see that a fixed layer delays the convergence time of participants.
In the Olivetti Faces experiments we see that in all cases all participants can achieve at least
95\% test set accuracy. All these results suggest that fixed layer can be used as an effective tool 
for increasing the embedding dimensionality without increasing the model complexity, with some penalty
mainly in the convergence speed. 

Unless otherwise stated, from now on, in all experiments, we share a fixed layer among the participants during their local trainings.

\subsection{Preventing GAN Attack}\label{sec:exp-preventing-gan-attack}

In this section we evaluate the success of our approach in preventing GAN attacks. 
We perform three sets of experiments to cover different aspects of our proposed solution:
\begin{enumerate}[label={\Roman*.}]
  \item\label{sec:exp-preventing-gan-attack-set1}
    The class key of one of the classes is given to the adversary.
  \item\label{sec:exp-preventing-gan-attack-set2}
    Adversary generates random keys that are $\delta$ far (measured in Euclidean distance)
    from some class key ($\lVert \psi_{\text{attack}} - \psi_c \rVert = \delta$ for some $c$).
  \item\label{sec:exp-preventing-gan-attack-set3}
    Adversary generates random keys.
\end{enumerate}
We evaluate samples generated by attackers both qualitatively (by visual inspection) and 
quantitatively by computing accuracies of the samples using an MNIST model pre-trained
on centralized data. For this evaluation, we pre-train a CNN model to
classify the MNIST digits over the complete dataset and use this model to 
classify synthetic samples generated by an adversary.

\begin{figure*}[t]
\centering
\begin{subfigure}[b]{\textwidth}
\centering
\begin{tabular}{c c c c c}
\includegraphics[width=0.10\linewidth]{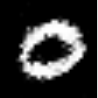} &
\includegraphics[width=0.10\linewidth]{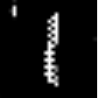} &
\includegraphics[width=0.10\linewidth]{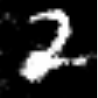} &
\includegraphics[width=0.10\linewidth]{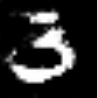} &
\includegraphics[width=0.10\linewidth]{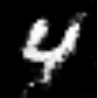} \\
\includegraphics[width=0.10\linewidth]{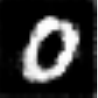} &
\includegraphics[width=0.10\linewidth]{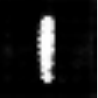} &
\includegraphics[width=0.10\linewidth]{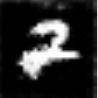} &
\includegraphics[width=0.10\linewidth]{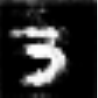} &
\includegraphics[width=0.10\linewidth]{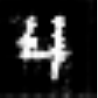}
\end{tabular}
\vspace{-0.2cm}
\caption{MNIST}
\end{subfigure}
\begin{subfigure}[b]{\textwidth}
\centering
\begin{tabular}{c c c c c}
\includegraphics[width=0.10\linewidth]{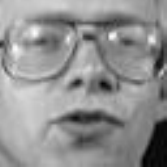} &
\includegraphics[width=0.10\linewidth]{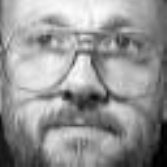} &
\includegraphics[width=0.10\linewidth]{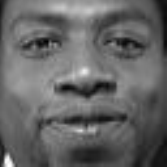} &
\includegraphics[width=0.10\linewidth]{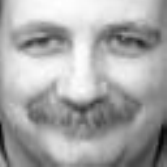} &
\includegraphics[width=0.10\linewidth]{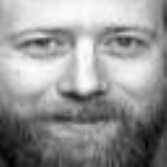} \\
\includegraphics[width=0.10\linewidth]{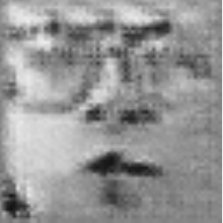} &
\includegraphics[width=0.10\linewidth]{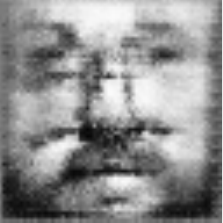} &
\includegraphics[width=0.10\linewidth]{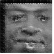} &
\includegraphics[width=0.10\linewidth]{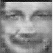} &
\includegraphics[width=0.10\linewidth]{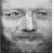} \\
\includegraphics[width=0.10\linewidth]{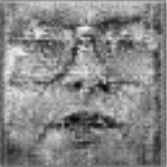} &
\includegraphics[width=0.10\linewidth]{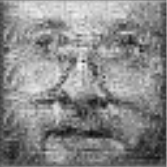} &
\includegraphics[width=0.10\linewidth]{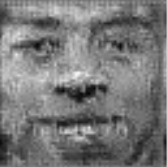} &
\includegraphics[width=0.10\linewidth]{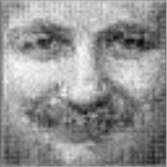} &
\includegraphics[width=0.10\linewidth]{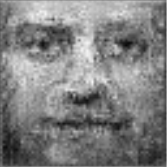} \\
\end{tabular}
\vspace{-0.2cm}
\caption{Olivetti Faces}
\end{subfigure}
\caption{Demonstrations of GAN attacks i) in the vanilla classification model where all class logits are seen by the participants~\cite{hitaj2017deep}, and 
ii) in our key-protected classification model when {\bf the attacker has access to the exact key of the class that it is attacking},
on the (a) MNIST and (b) Olivetti Faces datasets. 
We split the classes among two participants, one being an adversary.
In each dataset, we perform 5 different experiments, where the adversary attacks one of the classes owned by the victim. 
In MNIST (a), the victim owns the digit classes 0-4, and we present the GAN attack results in the vanilla (upper row) and our key-protected (lower row) classification models.
In Olivetti Faces (b), the victim owns the photos of 20 people, and, we show the original class samples (upper row),
the GAN attack results in the vanilla (middle row) and our key-protected (lower row) classification models.
Overall, the results demonstrate the effectiveness of the GAN attack, when key-based protection is not utilized.
}    
\label{fig:exp-attack-keys-given}
\end{figure*}

In Experiment~\ref{sec:exp-preventing-gan-attack-set1}, we demonstrate the extreme case in which adversary guesses the
same key of any class in collaborative learning framework.  Although this is very unlikely to happen in practice with
high-dimensional class keys, this case can be considered as a baseline where the GAN attack is expected to be
successful. The results for this case over the MNIST and Olivetti Faces datasets are presented in
Figure~\ref{fig:exp-attack-keys-given}. In the figure, we additionally show results when a vanilla classifier is being
used instead of our embedding based model. In these results, we observe that the attacker succeeds in
reconstructing target class images with high visual similarity both when using the vanilla classifier and the embedding
model with exact attack keys, as expected.  These results verify the effectiveness of the GAN attack in our experimental
setup.

\begin{figure*}[t]
  \centering
  \begin{adjustbox}{minipage=\linewidth,scale=.9}
  \centering
  \begin{tabular}{c c c c c c}
    \begin{subfigure}{0.10\textwidth}
        \centering
        \includegraphics[width=\linewidth]{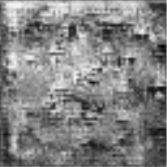}
        \caption{}
    \end{subfigure} &
    \begin{subfigure}{0.10\textwidth}
        \centering
        \includegraphics[width=\linewidth]{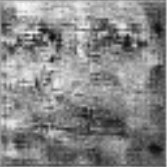}
        \caption{}
    \end{subfigure} &
    \begin{subfigure}{0.10\textwidth}
        \centering
        \includegraphics[width=\linewidth]{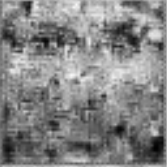}
        \caption{}
    \end{subfigure} &
    \begin{subfigure}{0.10\textwidth}
        \centering
        \includegraphics[width=\linewidth]{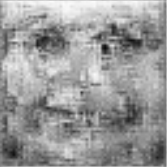}
        \caption{}
    \end{subfigure} &
    \begin{subfigure}{0.10\textwidth}
        \centering
        \includegraphics[width=\linewidth]{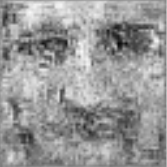}
        \caption{}
    \end{subfigure} &
    \begin{subfigure}{0.10\textwidth}
        \centering
        \includegraphics[width=\linewidth]{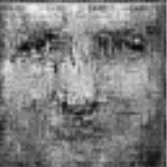}
        \caption{}
    \end{subfigure} \\
    \begin{subfigure}{0.10\textwidth}
        \centering
        \includegraphics[width=\linewidth]{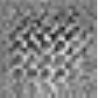}
        \caption{}
    \end{subfigure} &
    \begin{subfigure}{0.10\textwidth}
        \centering
        \includegraphics[width=\linewidth]{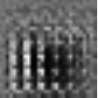}
        \caption{}
    \end{subfigure} &
    \begin{subfigure}{0.10\textwidth}
        \centering
        \includegraphics[width=\linewidth]{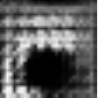}
        \caption{}
    \end{subfigure} &
    \begin{subfigure}{0.10\textwidth}
        \centering
        \includegraphics[width=\linewidth]{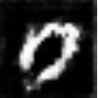}
        \caption{}
    \end{subfigure} &
    \begin{subfigure}{0.10\textwidth}
        \centering
        \includegraphics[width=\linewidth]{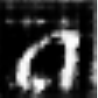}
        \caption{}
    \end{subfigure} &
    \begin{subfigure}{0.10\textwidth}
        \centering
        \includegraphics[width=\linewidth]{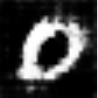}
        \caption{}
    \end{subfigure} 
  \end{tabular}
  \end{adjustbox}
  \caption{We approximate the maximum Euclidean distance between any class key and 
           $\psi(c_\text{attack})$ necessary for the adversary to succeed in attack. 
           From (a) to (f) and (g) to (l), reconstructions of the adversary when it generates random 
           keys that are $\delta \in \{1.3, 1.2, 1.1, 1.0, 0.5, 0.1 \}$ far from 
           the key of the class whose label is 24 in Olivetti Faces dataset and 0 in MNIST dataset, respectively.
           }
  \label{fig:exp-attack-delta}
\end{figure*}

In Experiment~\ref{sec:exp-preventing-gan-attack-set2}, we conduct a study to understand 
the success of the GAN attack as a function of the minimum similarity
between the GAN attack key and one of the actual class keys.
For this purpose, we provide the attacker a function that
takes the key of the victim class ($\psi_{\text{desired}}$), 
and a predefined degree of similarity ($\delta$),
and, generates an $\ell_2$ normalized random key $\psi_{\text{attack}}$ such that 
the generated key approximates the key of the victim class:
$ \lVert \psi_{\text{attack}} - \psi_{\text{desired}} \rVert \approx \delta$.
We experiment with several values of $\delta \in [0, 1.3]$ and see that
as $\delta$ decreases, the attacker starts to generate visually more similar images 
to the ones in the victim. By visually inspecting the reconstructions of the adversary, 
we deduce that for $\delta \ge 1.1$ on Olivetti Faces, the GAN attack produces incomprehensible results. 
In almost all attack experiments
we observe that generator of an adversary tends to collapse into
a single mode that is either a noise or a meaningful image, depending on 
the $\delta$ value. 
Consequently, on MNIST, we find a sharp threshold at $\delta = 0.5$ such that 
when $\delta \ge 0.5$ the synthesized samples become unrecognizable for the pre-trained MNIST classifier.
Reconstructions for different values of $\delta$ are given in Figure~\ref{fig:exp-attack-delta} for the Olivetti Faces and MNIST datasets.

\begin{figure*}
  \centering
  \begin{tabular}{c c c c c}
    \includegraphics[width=0.10\linewidth]{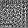} &
    \includegraphics[width=0.10\linewidth]{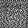} &
    \includegraphics[width=0.10\linewidth]{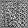} &
    \includegraphics[width=0.10\linewidth]{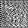} &
    \includegraphics[width=0.10\linewidth]{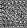} \\
  \end{tabular}
  \caption{GAN attack results on the MNIST dataset using random attack keys.
           We split MNIST among 5 participants which are also attackers. 
           Figures from left to right correspond to sample reconstructions obtained by each adversary, 
           when $d_{\text{key}} = 16834$.
           We run the experiments until each participant achieves 97\% accuracy on its
           local dataset. One can see that generators fail at capturing a valid digit 
           mode which indicates that the GAN attack is prevented.}
  \label{fig:exp-mnist-attack-random}
\end{figure*}

\begin{figure*}
  \centering
  \begin{tabular}{c c c c c}
    \includegraphics[width=0.10\linewidth]{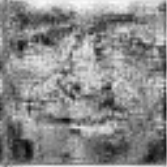} &
    \includegraphics[width=0.10\linewidth]{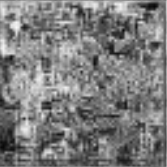} &
    \includegraphics[width=0.10\linewidth]{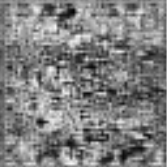} &
    \includegraphics[width=0.10\linewidth]{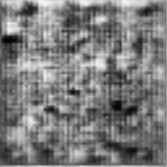}
  \end{tabular}
  \caption{GAN attack results on the Olivetti Faces dataset using random attack keys.
           From left to right, reconstructions of adversary for 
           $d_{\text{key}} \in \{128, 1024, 4096, 16834\}$, respectively.
           We show that the mode that GAN learns is likely to belong one of the classes in collaborative learning framework,
           for small $d_{\text{key}}$. However, as $d_{\text{key}}$ become larger,
           the quality of the GAN attack reconstructions degrade drastically. }
  \label{fig:exp-olivetti-attack-random}
\end{figure*}

In Experiment~\ref{sec:exp-preventing-gan-attack-set3}, we show that our model is robust against the 
GAN attack when there is no constraint on key generation, \ie~all keys are randomly and independently generated
by the participants, including the guessed ones. We show in 
Figure~\ref{fig:exp-mnist-attack-random} and  Figure~\ref{fig:exp-olivetti-attack-random} 
that for sufficiently large key dimensionalities,
the GAN attack fails on both MNIST and Olivetti Faces, respectively. 
In addition, we observe that the pre-trained MNIST classifier obtains $0\%$ accuracy on the generated examples,
which also support our claim that GAN attack fails. This situation
occurs in practice since the local generators of the adversaries collapse
to singular modes whose samples are unrecognizable for the pre-trained
classifier.

\subsection{Shared Classes Among Participants}\label{sec:sharedexp}

So far, we have assumed that there are no overlapping training classes across the participants. 
However, our approach can be utilized even in the cases where participants own samples 
from shared classes, without sharing their private class keys. We claim 
that as class keys becomes nearly orthonormal to each other,
a properly trained shared model would learn to map samples belonging to same class but hosted
by different participants to an space spanned by the private class keys
defined for that class by different participants. 
In this section, we show that our formulation works well 
when there exist shared classes among the participants.

Let $c$ be a class shared by the participants $i$ and $j$.
Let there exist two class key vectors in $\mathbb{R}^{2}$, namely $\psi_c^i$ and $\psi_c^j$,
which are two distinct keys of class $c$ generated by participants $i$ and $j$, respectively.
$\phi_{\theta}$ is the network that maps input data ($\eg$, images) into the embedding space.
Let $X_i^c$ and $X_j^c$ be the samples that participants $i$ and $j$ have 
for class $c$. For our analysis, we assume that the angle between $\psi_c^i$ and $\psi_c^j$ are
approximately orthogonal, which is correct in practice when the class keys are high-dimensional, $\ell_2$-normalized
vectors. In addition, as we expect observing similar examples in $X_i^c$ and $X_j^c$,
we assume, for simplicity, that the sets $X_i^c$ and $X_j^c$ 
contain a single shared sample $x^c$.

Then, since 
$\lVert \psi_c^i \rVert = \lVert \psi_c^j \rVert = 1$ by definition, and, %
 $\lVert \phi_{\theta}(x^c) \rVert = 1 $ due to the $\ell_2$-normalization layer at the output of the network,
the training objective, \ie~maximization of the dot product between the sample embedding and class key, 
for $x^c$ can equivalently expressed in terms of minimizing the angle between $\psi_c^i$ and $\phi_{\theta}(x^c)$ (denoted
by $\alpha_1$), and, between $\psi_c^j$ and $\phi_{\theta}(x^c)$ (denoted by $\alpha_2$)
~\footnote{$\langle \psi_c, \phi_{\theta}(x^c) \rangle = \lVert \psi_c \rVert_2 \cdot \lVert \phi_{\theta}(x^c) \rVert_2 \cdot \cos(\alpha) = \cos(\alpha)$ }.
In this simple example devised in $\mathbb{R}^2$, maximizing 
$ \langle \psi_c^i, \phi_{\theta}(x^c) \rangle $ and 
$ \langle \psi_c^j, \phi_{\theta}(x^c) \rangle $
with respect to  $\theta$ by participants $i$ and $j$ iteratively would converge to a model
such that 
$ \langle \psi_c^i, \phi_{\theta}(x^c) \rangle \approx %
  \langle \psi_c^j, \phi_{\theta}(x^c) \rangle \approx 0.7 $ (\ie~$\cos(\pi/4)$),
and $\alpha_1 \approx \alpha_2 \approx \pi/4 $.

\begin{figure}
\begin{subfigure}{0.49\linewidth}
\centering
\includegraphics[width=0.95\linewidth]{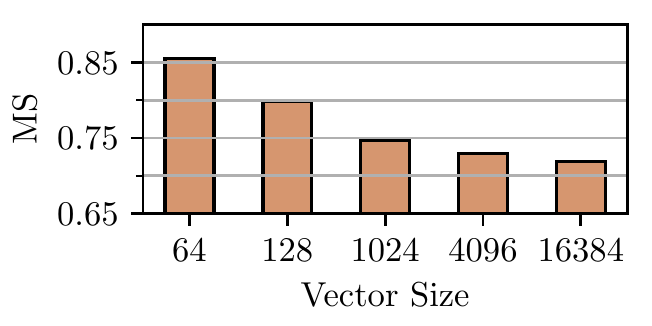}
\end{subfigure}
\begin{subfigure}{0.49\linewidth}
\centering \includegraphics[width=0.95\linewidth]{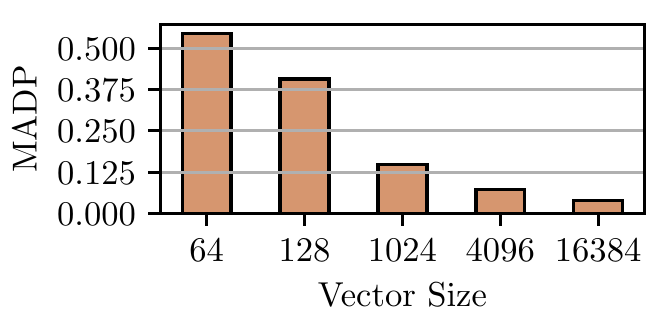}
\end{subfigure}
\caption{(Left) We randomly generate $\ell_2$ normalized $ \psi_c^i,\psi_c^j,\phi(x^c) \in
    \mathbb{R}^{d_{\text{key}}} $ for $d_{\text{key}} \in \{ 64, 128, 1024, 4096, 16384 \} $. Then we find the optimal
    $\phi(x^c)$ such that its dot product with $\psi_c^i$ and $\psi_c^j$ are maximized. Bars indicate that as we
    increase the dimensionality, $\psi_c^i$ and $\psi_c^j$ are more likely to be orthogonal, therefore $\phi(x^c)$ ends
    up being a vector at $0.7$ correlation with each one. MS: Maximum score.  (Right) We generate new $\ell_2$
    normalized vectors in $\mathbb{R}^{d_{\text{key}}}$ for $d_{\text{key}} \in \{ 64, 128, 1024, 4096, 16384 \} $, and
    check their maximum absolute dot product with the final $\phi(x^c)$. We see that as we increase $d_{\text{key}}$,
    $\phi(x^c)$ converges towards the space spanned by $\psi_c^i$ and $\psi_c^j$. This confirms that the approach is
    likely to behave well when training over shared classes, despite using different private keys across the
    participants. MADP: Maximum absolute dot product.  }\label{fig:exp-class-disjointness}
\end{figure}

In consideration of this behavior in much higher dimensional spaces, we perform the 
following experiments. For $d_{\text{emb}}=\{64, 256, 1024, 4096, 16384\}$, we
generate three $\ell_2$ normalized vectors, namely $\psi_c^i$, $\psi_c^j$ and $\phi(x^c)$.
Then we update $\phi(x^c)$ to maximize 
$ \langle \psi_c^i, \phi(x^c) \rangle $ and 
$ \langle \psi_c^j, \phi(x^c) \rangle $ until convergence, by simple gradient descent.
We repeat this procedure 1000 times and plot maximum of the scores obtained either by
$ \langle \psi_c^i, \phi(x^c) \rangle $ or by
$ \langle \psi_c^j, \phi(x^c) \rangle $ for all $d_{\text{key}}$
in Figure~\ref{fig:exp-class-disjointness} (left). We see that scores converge to 0.7 as we increase
the class key dimensionality.

We continue our analysis by measuring the dot product of the tuned $\phi(x^c)$ with 
random class keys representing other classes. Our purpose is to interpret the optimization output
of the previous experiment. For each tuned $\phi(x^c)$ we generate $\ell_2$ normalized
vectors $\psi^{\text{new}}_k$ for $k=1,\ldots,1000$ and plot maximum absolute dot 
products of all $\psi^{\text{new}}_k$ and $\phi(x^c)$,
in Figure~\ref{fig:exp-class-disjointness} (right). Results indicate that when 
$d_{\text{key}}$ is sufficiently high, multiple participants can declare the same set of labels. For each
common label, the accumulated model is likely to map samples belonging to the shared class
onto a space spanned by the embeddings defined for that class
among the participants. The resulting space is likely to be nearly orthogonal to any other class key.
At test time, as labels are assigned according to the Equation~\ref{eq:prediction},
having multiple keys for a class does not constitute an issue, \ie~we can simply assign 
a test sample to the class whose one of the keys maximizes the classification score.

These observations confirm that the approach is likely to behave well when training over shared classes,
despite using different private keys across the participants. Overall, the network is likely to
map samples to points that are highly correlated with all duplicate keys of their ground-truth classes, and highly 
uncorrelated with the other ones. In fact, we have empirically 
verified that training with shared classes perform with no observable issues on both MNIST and Olivetti Faces
(example results are omitted for brevity).
Our experiments also suggest that 
when using sufficiently {\em long} (\ie~high-dimensional) keys,
generator of each attacker is able to 
capture some mode that does not match with any of the modes of the classes in the collaborative learning framework,
according to fake class key generated by the attacker.

\subsection{Evaluating the Key Based Regression Loss}\label{sec:key-versus-xent}

We are also curious about the performance of our regression based objective that we
formulate in Equation~\ref{eq:trainobj3} when it is used out of the privacy context,
\eg~in the supervised classification setup where all the training data is centralized.
For this purpose, we train several ResNet-10~\cite{He_2016_CVPR} models on the CIFAR-10 and CIFAR-100
datasets~\cite{krizhevsky2009learning}
by optimizing the cross entropy loss and the key-based cosine similarity loss, which we refer to as the {\em key-based
regression}, and compare their results. 

In the original ResNet models, there is one fully connected layer mapping non-negative image embeddings
(obtained at the output of the last convolutional layer) to classification scores by computing
dot products between weight vectors in the fully connected layer and the image embeddings
then adding bias terms in the fully connected layer.
Instead, in our experiments we compute cosine similarity scores between the weight vectors 
in the fully connected layer and the image embeddings.
In \texttt{cross entropy} experiments, the weights of the fully connected layer are trainable,
whereas in \texttt{key-based regression} experiments we replace the trainable weights with
fixed orthonormal class keys, which we obtain by QR decomposition before training starts.
Finally, we modify the last convolutional layer by replacing the ReLU activation
function with the $\tanh$, based on the observation that computing cosine similarities by using non-negative image embeddings $\phi_\theta(x)$ 
tends to make training unstable.

\begin{table}
\centering
\begin{tabular}{c c c}
    \textbf{Objective} & \textbf{Dataset} & \textbf{Top-1} \\ \toprule  
    Key-based regression & CIFAR-10 & $\mathbf{94.8}$\% \\  \midrule
    Cross entropy & CIFAR-10 & $94.0$\% \\  \midrule
    Key-based regression & CIFAR-100 & $73.6$\% \\  \midrule
    Cross entropy & CIFAR-100 & $\mathbf{74.2}$\% \\  \midrule
\end{tabular}
\caption{Key-based regression versus cross entropy.}
\label{tab:key-loss-vs-xent}
\end{table}

Top-1 accuracies obtained for both loss functions are shown in Table~\ref{tab:key-loss-vs-xent}.
We see that \texttt{key-based regression} outperforms cross entropy on CIFAR-10 by a small margin,
whereas on \texttt{cross-entropy} performs better on CIFAR-100.
These results indicate that orthonormal class keys
enforce the network to produce sufficiently orthonormal {\em class codes} at the 
last fully connected layer, and make the overall training formulation discriminative.
We leave an elaborate study for key-based regression in a future work.

\section{Conclusion}\label{sec:conclusion}
We have presented a collaborative learning technique that is resilient to the GAN attack.
More specifically, we have introduced a classification model for participants, where random class keys represents classes.
This key-based model provides effective learning over the participants and by utilizing high dimensional keys, class scores of an input is protected against an active adversary that may aim to execute a GAN attack.
We have presented a principled training technique for the case of restricting the access of each participant to only its own private keys, and, a way to increase key dimensionality without  increasing the model complexity.
We have verified the effectiveness of our formulation by empirically showing that (i) the adversary is no longer able to choose which class to exploit, and (ii) generator trained by the adversary cannot capture data distribution well enough to reconstruct any class.
We believe that the proposed approach makes a step towards making collaborative learning safe and practical, which can potentially have a fundamental impact on learning models in sensitive data domains.

\section{Acknowledgements}
This work was supported in part by the TUBITAK Grant 116E445.

\bibliographystyle{elsarticle-num}
\bibliography{paper}

\clearpage
\onecolumn
\appendix
\section{Softmax Over Infinitely-many Classes}\label{appx:exppsi}

In this section, we show the detailed derivation of our softmax generalization to infinitely-many classes. Below, we use the subscript notation on $\psi$ to denote dimensions, instead of classes, and, use $d$ instead of $d_\text{key}$ for brevity. 
In order to compute $p(c|x) = \frac{\exp g_\theta(x,\psi_c)}{\mathbb{E}_{\psi}\left[\exp( g_\theta(x,\psi) ) \right]}$, we need to compute the 
expectation in the denominator:
\begin{eqnarray}
\mathbb{E}_{\psi \sim \mathcal{N}} \left[ \exp( g_\theta(x,\psi) ) \right] = \int_{\psi} f(\psi) \exp \left( g_\theta(x,\psi) \right) d\psi .
\end{eqnarray}
Since the class keys are assumed to follow multivariate standard normal distribution, the integration can be simplified as follows:
\begin{eqnarray}
     \int_{\psi} f(\psi) \exp \left( g_\theta(x,\psi) \right) d\psi = \int_{\psi_1} ... \int_{\psi_d} f(\psi_1) ... f(\psi_d) \exp\left( g(x,\psi) \right) d\psi_1 ...
     d\psi_d .
\end{eqnarray}
By re-arranging the terms and plugging-in Eq.~\ref{eq:scoringfunc}, we obtain:
\begin{eqnarray}
    \int_{\psi_d} f(\psi_d) ... \int_{\psi_1}  f(\psi_1) \exp\left( {\psi}^T \phi(x) \right) d\psi_1 ... d\psi_d ,
\end{eqnarray}
By converting the exponent of summation into a product of exponents, we obtain:
\begin{eqnarray}
    \int_{\psi_d} f(\psi_d) \exp\left({\phi_d(x) \psi_d}\right) ... \int_{\psi_1} f(\psi_1) \exp\left({\phi_1(x) \psi_1}\right) d\psi_1 ... d\psi_d , \label{eq:intwithalpa1}
\end{eqnarray}
where $\phi_i(x)$ refers to the $i$-th dimension of the $\phi(x)$ vector.
Here, the inner-most term can be integrated out easily:
\begin{eqnarray}
    &~& \int_{\psi_1}  f(\psi_1) \exp\left({\phi_1(x) \psi_1}\right) d\psi_1   \\
    &=& \frac{1}{\sqrt{2\pi}} \int_{\psi_1}  \exp\left({-\frac{1}{2}\psi_1^2}\right) \exp\left({\phi_1(x) \psi_1}\right) d\psi_1 \\
    &=& \frac{1}{\sqrt{2\pi}} \int_{\psi_1}  \exp\left({-\frac{1}{2} (\psi_1 - \phi_1(x))^2}\right) \exp\left({\frac{1}{2}\phi_1(x)^2}\right) d\psi_1 \\
    &=& \exp\left({\frac{1}{2}\phi_1(x)^2}\right),
\end{eqnarray}
where the last step is based on the observation that the remaining terms correspond to integrating out Gaussian probability density function with $\mu=\phi_1(x)$ and $\sigma=1$. Now, we can 
plug-in this result into Eq.~\ref{eq:intwithalpa1} and repeatedly integrate out each dimension:
\begin{eqnarray}
  \exp\left({\frac{1}{2}\phi_1(x)^2}\right) \int_{\psi_d} f(\psi_d) \exp\left({\phi_d(x) \psi_d}\right) ... \int_{\psi_{2}} f(\psi_{2}) \exp\left({\phi_{2}(x) \psi_{2}}\right) d\psi_2 ... d\psi_d \label{eq:intwithalpa3} \\
 = \prod_{i=1}^{d} \exp\left({\frac{1}{2}\phi_i(x)^2}\right) = \exp\left( \frac{1}{2} \sum_{i=1}^{d} \phi_i(x)^2 \right) = \exp\left( \frac{1}{2} \|\phi(x)\|^2 \right).
\end{eqnarray}
Finally, by plugging-in Eq.~\ref{eq:phidef}, we obtain a constant value of $\exp(0.5)$:
\begin{eqnarray}
    \mathbb{E}_{\psi \sim \mathcal{N}} \left[ \exp( g_\theta(x,\psi) ) \right] &=& \exp\left( \frac{1}{2} \|\phi(x)\|^2  \right) \\
    &=& \exp\left( \frac{1}{2} \left\| \frac{\phi_u(x)}{\|\phi_u(x)\|}\right\|^2  \right) \\
    &=& \exp\left( 0.5 \right),
\end{eqnarray}
which completes the proof that our generalization of softmax to infinitely-many classes takes the following final form:
\begin{eqnarray}
    \frac{\exp g_\theta(x,\psi_c)}{\mathbb{E}_{\psi}[\exp g_\theta(x,\psi) ]} = \frac{1}{\exp(0.5)} \exp\left( g_\theta(x,\psi_c) \right) .
\end{eqnarray}


\end{document}